\documentclass{article}

\PassOptionsToPackage{numbers,compress}{natbib}

\usepackage[preprint]{tackling_climate_workshop_style}

\usepackage[utf8]{inputenc} 
\usepackage[T1]{fontenc}    
\usepackage{hyperref}       
\usepackage{url}            
\usepackage{booktabs}       
\usepackage{amsfonts}       
\usepackage{nicefrac}       
\usepackage{microtype}      

\usepackage{amsmath, amssymb}   
\usepackage{graphicx}           
\usepackage{subcaption}         
\usepackage{float}              
\usepackage{wrapfig}            
\usepackage{xcolor}             
\usepackage{rotating}
\usepackage{enumitem}           
\usepackage{tikz}               
\usetikzlibrary{positioning, shapes.geometric, arrows.meta, shadows}
\usepackage{pgfplots}           
\pgfplotsset{compat=1.18}
\definecolor{cUnet}{HTML}{2A78D6}   
\definecolor{cINR-SR}{HTML}{008300}   
\definecolor{cSE-OT}{HTML}{E87BA4}   
\definecolor{cCLR}{HTML}{EDA100}    
\definecolor{cInterp}{HTML}{1BAF7A} 
\pgfplotsset{
  benchbar/.style={
    ybar=0.5pt, bar width=3.4pt,
    width=\linewidth, height=4.4cm,
    symbolic x coords={MAE,SSIM,Wasserstein,Spectral,Gradient},
    xtick=data,
    xticklabels={MAE,SSIM,Wass.,Spec.,Grad.},
    x tick label style={font=\scriptsize, rotate=25, anchor=east},
    ymin=0, ymax=1.18, ytick={0,0.5,1},
    tick label style={font=\scriptsize},
    ylabel={score}, ylabel style={font=\scriptsize},
    ymajorgrids=true, grid style={gray!20}, axis line style={gray!55},
    title style={font=\small}, enlarge x limits=0.16,
  },
}
\pgfplotsset{ benchbarkr/.style={benchbar, ymax=1.02, enlarge x limits=0.14,
    x tick label style={font=\scriptsize, rotate=40, anchor=east}} }
\usepgfplotslibrary{groupplots}  

\usepackage{booktabs}           
\usepackage{longtable}
\usepackage{multirow}           
\usepackage[ruled,vlined]{algorithm2e} 
\usepackage{adjustbox}
\usepackage{xfp} 

\usepackage[normalem]{ulem}     
\usepackage{numprint}
\usepackage{siunitx}
\usepackage{csquotes}           

\title{Beyond Pointwise Error: A Multi-Metric Evaluation of Spatial Climate Downscaling}

\author{%
   Loys Masquelier\\
   EDF R\&D, Palaiseau, France\\
   \texttt{loys.masquelier@edf.fr} \\
   \And
   Etienne Le Naour \\
   EDF R\&D, Palaiseau, France\\
   \texttt{etienne.le-naour@edf.fr} \\
 }

\begin{document}

\maketitle


\begin{abstract}
Climate downscaling aims to reconstruct fine scale spatial fields from coarse resolution inputs. Evaluating the quality of these reconstructions is challenging: low pointwise error can come at the cost of fine scale variability, while realistic spatial variability can be achieved with inaccurate local structures. The evaluation metric can therefore change which method appears to perform best. This work presents a multi metric benchmark comparing five spatial downscaling methods on ERA5 temperature, wind, and precipitation fields. Five criteria assess complementary properties: pointwise error, structural similarity, distribution error, spectral error, and gradient error. The results reveal a systematic trade off between spatial fidelity and fine scale variability. Some methods perform best on pointwise and spatially aligned metrics, but lose high frequency content, while others preserve substantially more spectral variability at the cost of less accurately positioned local structures. Consequently, method rankings change across metrics and variables. These results show that there is no single best downscaling method. Multi metric evaluation is therefore essential for assessing which properties of a climate field are preserved.
\end{abstract}

\section{Introduction}

Climate adaptation increasingly requires information at spatial scales finer than those resolved by global and regional climate models. Downscaling methods bridge this gap by reconstructing fine-scale fields from coarse-resolution inputs. Classical approaches include dynamical downscaling \citep{hong2014dynamical, xu2019dynamical} and statistical methods \citep{benestad2008empirical, wilby2013statistical}, while recent work increasingly relies on machine learning and super-resolution~\citep{Michelangeli2009-zb,Robin2019-rz,dong2015imagesuperresolutionusingdeep,kim2016accurateimagesuperresolutionusing,ronneberger2015unetconvolutionalnetworksbiomedical,ledig2017photorealisticsingleimagesuperresolution}.

However evaluating a downscaled climate field is more subtle than evaluating a conventional image \citep{zeraatkar2026frequency}. A reconstruction can have low pointwise error while being overly smooth and losing fine-scale variability \citep{freirich2021theory, blau2018perception}. Conversely, it can preserve realistic spatial variability while placing individual structures at the wrong locations \citep{roberts2008scale}. These trade-offs also depend on the meteorological variable, from smooth temperature fields to turbulent winds and sparse, intermittent precipitation. Different metrics therefore reward different properties of a reconstruction.

This work investigates how the choice of evaluation metric changes which downscaling method appears to perform best. To this end, five spatial downscaling methods are benchmarked on \textit{ERA5} \texttt{temperature}, \texttt{wind}, and \texttt{precipitation} fields over France~\citep{era5,era5_1940}. The benchmark includes four learned approaches, \texttt{residual U-Net}, \texttt{INR-SR}, \texttt{SE-OT}, and \texttt{CLR}, together with \texttt{bilinear interpolation}. Reconstructions are evaluated using five complementary criteria covering pointwise error, local structure, value distribution, spatial frequency content, and fine-structure localisation.

The results reveal a systematic trade-off: \texttt{U-Net} performs best on pointwise and spatially aligned metrics but produces smoother fields with less high-frequency content, whereas \texttt{SE-OT} and \texttt{CLR} preserve substantially more spectral variability at the cost of less accurate fine-structure localisation. Consequently, method rankings vary across metrics and meteorological variables, showing that a single evaluation criterion can hide important differences between downscaling methods.

These findings motivate a broader evaluation framework that captures complementary aspects of reconstruction quality. The main contributions of this work are: 

$\bullet$ \textbf{A multi-metric benchmark:} a benchmark spanning complementary properties of downscaled climate fields, including pointwise fidelity, local structure, value distribution, spectral content, and fine-scale localisation. 

$\bullet$ \textbf{A metric-dependent ranking:} evidence that method rankings can reverse across evaluation criteria, with the magnitude of these reversals growing with the spatial complexity of the meteorological variable.

\section{Benchmark Construction}
\label{sec:benchmark}

The benchmark is designed to compare (i) downscaling methods with different reconstruction mechanisms and to evaluate the (ii)  properties they preserve across (iii) meteorological variables of varying spatial complexity. For more details about the methods please refer to Appendix~\ref{app:methods}.


\paragraph{(i) Methods.}
In this work, five spatial downscaling methods are compared, including four learned approaches and \texttt{bilinear interpolation}. They span different reconstruction mechanisms and therefore different compromises between \textit{spatial fidelity}, \textit{statistical realism}, and \textit{fine-scale variability}. \texttt{Residual U-Net}~\citep{Zhang_2018} predicts a residual correction to a bicubic-interpolated field; its convolutional architecture and point-to-point objective favour spatially aligned reconstructions. \texttt{Implicit Neural Representation Super-Resolution (INR-SR)} uses an implicit neural representation with periodic SIREN activations~\citep{sitzmann2020implicitneuralrepresentationsperiodic}, whose latent code, inferred from the low-resolution input, drives a continuous high-resolution reconstruction \citep{NEURIPS2023_df543023, naour2024timeseriescontinuousmodeling}. \texttt{Sparse Entropic Optimal Transport (SE-OT)} reconstructs a field as a weighted barycentre of high-resolution reference examples, with weights obtained from a sparse, entropy-regularised optimal-transport formulation~\citep{cuturi2013sinkhorndistanceslightspeedcomputation,peyré2020computationaloptimaltransport}. \texttt{Contrastive Latent Retrieval (CLR)} learns a latent representation of low-resolution fields by contrastive learning~\citep{chen2020simpleframeworkcontrastivelearning} and retrieves similar references, combining their high-resolution fields with similarity-based weights. \texttt{Bilinear interpolation} provides a simple non-learned baseline for comparison.

\paragraph{(ii) Metrics.}
Five complementary criteria are used to assess different properties of the reconstructions (formal definitions in Appendix~\ref{app:metrics}). The \texttt{Mean Absolute Error} (\texttt{MAE})~\citep{Willmott2005} measures pointwise fidelity, while \texttt{Structural Similarity Index Measure} (\texttt{SSIM}) ~\citep{ssim} evaluates local structural similarity. The \texttt{Wasserstein distance}~\citep{arjovsky2017wassersteingan,optimtransport} compares the distributions of predicted and reference values independently of their spatial locations. The \texttt{spectral error}~\citep{Harris_2022} evaluates whether spatial variability is distributed correctly across frequencies. Finally, the \texttt{gradient error}~\citep{gradient} measures fine-scale structure while retaining spatial alignment: a structure contributes positively only when it is reproduced at the correct location.

\paragraph{(iii) Data.}
The benchmark is conducted on daily \textit{ERA5} reanalysis data from 1940--2020, centred on France, for four variables: near-surface temperature (\texttt{tas}), zonal and meridional wind (\texttt{u10}, \texttt{v10}), and precipitation (\texttt{pr}). High-resolution fields have size $61\times42$, yielding \numprint{29586} examples with an 80/10/10 train/validation/test split; low-resolution fields ($7\times9$) are obtained synthetically by bilinear downsampling. These variables cover distinct levels of spatial complexity: temperature is predominantly smooth and low-frequency, wind contains stronger small-scale variability, and precipitation is sparse and intermittent.

\section{Key Results}
\label{sec}

\begin{wrapfigure}[37]{r}{0.36\linewidth}
  \centering
  \includegraphics[width=\linewidth]{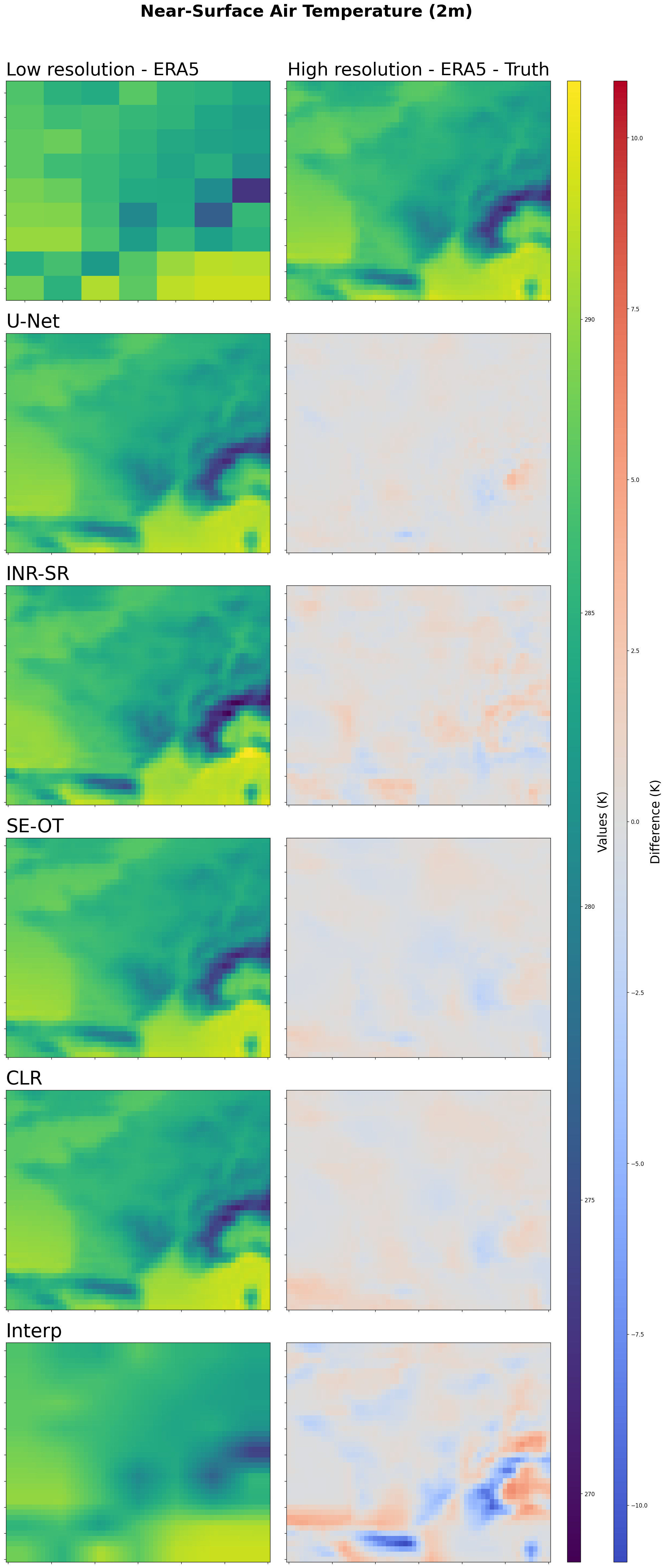}
  \caption{Reconstructions of the same \texttt{tas} field from a common coarse-resolution input. The methods trade off spatial fidelity against fine-scale variability.}
  \label{fig:first_page_tas}
\end{wrapfigure}

\begin{figure}[t]
  \centering
  \pgfplotsset{row2/.style={
      axis x line*=top,
      every x tick label/.style={rotate=40, anchor=center, font=\scriptsize,
      xshift=5pt, yshift=7pt},
      xlabel style={at={(axis description cs:0.5,0)}, anchor=north,
      font=\scriptsize, yshift=-1pt}}}
  \begin{tikzpicture}
  \begin{groupplot}[
      group style={group size=3 by 2, horizontal sep=6pt, vertical sep=24pt,
                   y descriptions at=edge left},
      benchbarkr, scale only axis, width=0.29\textwidth, height=2.4cm,
      axis x line*=bottom, axis y line*=left,
  ]

    \nextgroupplot[title={\texttt{tas}}, xticklabels={},
        title style={font=\scriptsize, yshift=-6pt}]
      \addplot[fill=cUnet,draw=cUnet] coordinates
      {(MAE,1.00) (SSIM,1.00) (Wasserstein,1.00) (Spectral,0.49) (Gradient,1.00)};
      \addplot[fill=cINR-SR,draw=cINR-SR] coordinates
      {(MAE,0.36) (SSIM,0.24) (Wasserstein,0.17) (Spectral,0.43) (Gradient,0.42)};
      \addplot[fill=cSE-OT,draw=cSE-OT] coordinates
      {(MAE,0.49) (SSIM,0.64) (Wasserstein,0.31) (Spectral,1.00) (Gradient,0.77)};
      \addplot[fill=cCLR,draw=cCLR] coordinates
      {(MAE,0.43) (SSIM,0.66) (Wasserstein,0.22) (Spectral,0.98) (Gradient,0.75)};
      \addplot[fill=cInterp,draw=cInterp] coordinates
      {(MAE,0.25) (SSIM,0.08) (Wasserstein,0.25) (Spectral,0.13) (Gradient,0.28)};

    \nextgroupplot[title={\texttt{pr}}, xticklabels={},
        title style={font=\scriptsize, yshift=-6pt}]
      \addplot[fill=cUnet,draw=cUnet] coordinates
      {(MAE,1.00) (SSIM,1.00) (Wasserstein,1.00) (Spectral,0.45) (Gradient,1.00)};
      \addplot[fill=cINR-SR,draw=cINR-SR] coordinates
      {(MAE,0.78) (SSIM,0.72) (Wasserstein,0.49) (Spectral,0.44) (Gradient,0.82)};
      \addplot[fill=cSE-OT,draw=cSE-OT] coordinates
      {(MAE,0.60) (SSIM,0.45) (Wasserstein,0.42) (Spectral,1.00) (Gradient,0.78)};
      \addplot[fill=cCLR,draw=cCLR] coordinates
      {(MAE,0.58) (SSIM,0.46) (Wasserstein,0.34) (Spectral,0.95) (Gradient,0.83)};
      \addplot[fill=cInterp,draw=cInterp] coordinates
      {(MAE,0.77) (SSIM,0.67) (Wasserstein,0.61) (Spectral,0.28) (Gradient,0.84)};

    \nextgroupplot[
        axis y line=none, axis x line=none, ymin=0, ymax=1.02,
        legend columns=3,
        legend style={at={(0.5,0.5)}, anchor=center, draw=gray!40,
        font=\scriptsize, /tikz/every even column/.append style={column sep=5pt}}]
      \addplot[draw=none, forget plot] coordinates
      {(MAE,0) (SSIM,0) (Wasserstein,0) (Spectral,0) (Gradient,0)};
      \addlegendimage{fill=cUnet,draw=cUnet}
      \addlegendentry{U-Net}
      \addlegendimage{fill=cINR-SR,draw=cINR-SR}
      \addlegendentry{INR-SR}
      \addlegendimage{fill=cSE-OT,draw=cSE-OT}
      \addlegendentry{SE-OT}
      \addlegendimage{fill=cCLR,draw=cCLR}
      \addlegendentry{CLR}
      \addlegendimage{fill=cInterp,draw=cInterp}
      \addlegendentry{Interp}

    \nextgroupplot[row2, xlabel={\texttt{u10}}]
      \addplot[fill=cUnet,draw=cUnet] coordinates
      {(MAE,1.00) (SSIM,1.00) (Wasserstein,1.00) (Spectral,0.75) (Gradient,1.00)};
      \addplot[fill=cINR-SR,draw=cINR-SR] coordinates
      {(MAE,0.61) (SSIM,0.47) (Wasserstein,0.37) (Spectral,0.76) (Gradient,0.70)};
      \addplot[fill=cSE-OT,draw=cSE-OT] coordinates
      {(MAE,0.42) (SSIM,0.31) (Wasserstein,0.23) (Spectral,0.98) (Gradient,0.62)};
      \addplot[fill=cCLR,draw=cCLR] coordinates
      {(MAE,0.40) (SSIM,0.32) (Wasserstein,0.18) (Spectral,1.00) (Gradient,0.63)};
      \addplot[fill=cInterp,draw=cInterp] coordinates
      {(MAE,0.34) (SSIM,0.22) (Wasserstein,0.29) (Spectral,0.25) (Gradient,0.39)};

    \nextgroupplot[row2, xlabel={\texttt{v10}}]
      \addplot[fill=cUnet,draw=cUnet] coordinates
      {(MAE,1.00) (SSIM,1.00) (Wasserstein,1.00) (Spectral,0.69) (Gradient,1.00)};
      \addplot[fill=cINR-SR,draw=cINR-SR] coordinates
      {(MAE,0.63) (SSIM,0.52) (Wasserstein,0.39) (Spectral,0.62) (Gradient,0.73)};
      \addplot[fill=cSE-OT,draw=cSE-OT] coordinates
      {(MAE,0.44) (SSIM,0.32) (Wasserstein,0.25) (Spectral,1.00) (Gradient,0.66)};
      \addplot[fill=cCLR,draw=cCLR] coordinates
      {(MAE,0.42) (SSIM,0.34) (Wasserstein,0.19) (Spectral,0.96) (Gradient,0.67)};
      \addplot[fill=cInterp,draw=cInterp] coordinates
      {(MAE,0.38) (SSIM,0.25) (Wasserstein,0.31) (Spectral,0.26) (Gradient,0.45)};

    \nextgroupplot[row2, xlabel={\texttt{u10-v10}}]
      \addplot[fill=cUnet,draw=cUnet] coordinates
      {(MAE,1.00) (SSIM,1.00) (Wasserstein,1.00) (Spectral,0.73) (Gradient,1.00)};
      \addplot[fill=cINR-SR,draw=cINR-SR] coordinates
      {(MAE,0.66) (SSIM,0.51) (Wasserstein,0.47) (Spectral,0.69) (Gradient,0.74)};
      \addplot[fill=cSE-OT,draw=cSE-OT] coordinates
      {(MAE,0.36) (SSIM,0.25) (Wasserstein,0.19) (Spectral,0.99) (Gradient,0.61)};
      \addplot[fill=cCLR,draw=cCLR] coordinates
      {(MAE,0.35) (SSIM,0.26) (Wasserstein,0.17) (Spectral,1.00) (Gradient,0.62)};
      \addplot[fill=cInterp,draw=cInterp] coordinates
      {(MAE,0.34) (SSIM,0.20) (Wasserstein,0.30) (Spectral,0.26) (Gradient,0.40)};

  \end{groupplot}
  \end{tikzpicture}
  \caption{Performance on each metric, normalised score (``best $=1$'').
  The ranking of downscaling methods depends strongly on the property being
  evaluated. \texttt{U-Net} dominates spatially aligned metrics, whereas \texttt{SE-OT} and
  \texttt{CLR} dominate spectral fidelity.}
  \label{fig:bench_grid}
\end{figure}
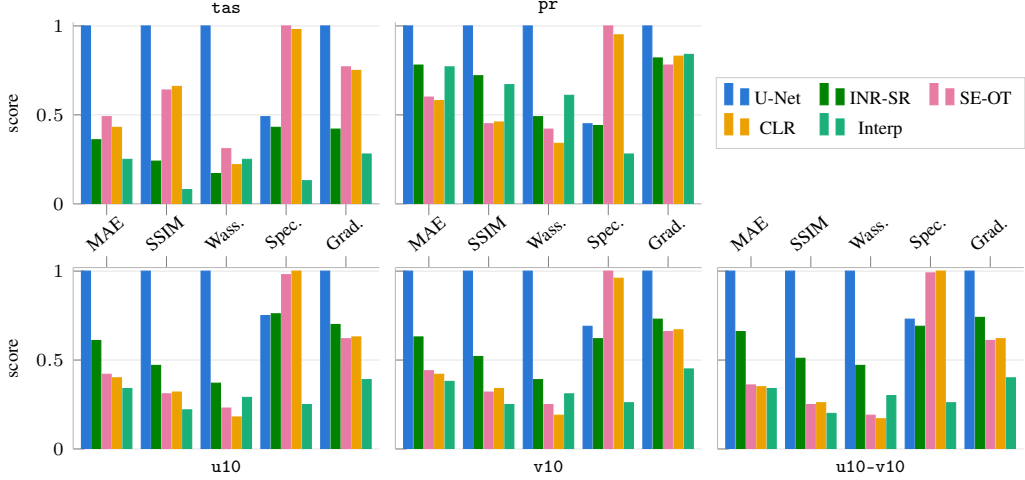

Evaluating a downscaled climate field is inherently multi-faceted: methods can produce qualitatively different reconstructions from the same coarse-resolution input, without a single visually or numerically obvious winner. Figure~\ref{fig:first_page_tas} illustrates this point for \texttt{tas}. \texttt{U-Net} produces a smooth, well-aligned reconstruction, whereas the retrieval-based methods preserve more small-scale variability; \texttt{INR-SR} also recovers fine-scale structures, but with a different spatial organisation. A sharper reconstruction is not necessarily more accurate, while a smoother field may achieve lower pointwise error by avoiding uncertain small-scale structures. Figure~\ref{fig:bench_grid} quantifies this trade-off across the four variables and five metrics, with scores normalised independently for each variable and metric (best $=1$). Two key patterns emerge: (i) method rankings depend strongly on the property being evaluated, and (ii) the magnitude of this trade-off varies substantially across meteorological variables.

\noindent\textbf{(i) Metric choice can reverse the ranking.}
Across all four variables, \texttt{U-Net} dominates the spatially aligned metrics (\texttt{MAE}, \texttt{SSIM}, \texttt{Wasserstein}, and \texttt{gradient error}), whereas \texttt{SE-OT} and \texttt{CLR} dominate the \texttt{spectral score}, reaching scores of $0.95$--$1.00$, compared with $0.45$--$0.75$ for \texttt{U-Net} and $0.13$--$0.28$ for interpolation. The contrast is clearest for \texttt{tas}: \texttt{U-Net} ranks first on four of five metrics, while \texttt{SE-OT} achieves the best \texttt{spectral score}, closely followed by \texttt{CLR}. Thus, a benchmark dominated by pointwise and spatially aligned criteria would select \texttt{U-Net}, whereas one emphasising spatial-frequency variability would select \texttt{SE-OT} or \texttt{CLR}. This reversal follows directly from the reconstruction mechanisms. The point-to-point objective of \texttt{U-Net} favours conservative, target-aligned predictions but tends to attenuate uncertain high-frequency structures. By contrast, \texttt{SE-OT} and \texttt{CLR} retrieve high-resolution reference examples that preserve realistic spatial variability, at the cost of potentially misplacing fine-scale structures when the retrieved references are only approximately similar to the target. \texttt{INR-SR}, through its periodic SIREN representation, also recovers fine-scale variability and is particularly competitive for wind and precipitation. The metrics therefore capture different notions of fidelity: \texttt{MAE}, \texttt{SSIM}, \texttt{Wasserstein}, and \texttt{gradient error} reward agreement with the target field, whereas the \texttt{spectral score} rewards preservation of variability across spatial scales. The ``best'' downscaler is therefore not an intrinsic property of the model, but a consequence of what the benchmark chooses to reward.

\noindent\textbf{(ii) Reconstruction difficulty varies systematically across meteorological variables.} Table~\ref{tab:rough_mae} shows that spatial roughness increases from \texttt{tas} to wind to precipitation, approximately following $\texttt{tas} < \texttt{u10} \simeq \texttt{v10} < \texttt{pr}$. The \texttt{MAE} of \texttt{U-Net} follows the same ordering, indicating that reconstruction difficulty is strongly associated with the amount of small-scale variability in the target field. \texttt{tas} is predominantly smooth and low-frequency and is therefore the easiest variable to reconstruct; \texttt{wind} exhibits stronger small-scale variability, while precipitation is sparse and intermittent, making it substantially more challenging.

\begin{table}[h!]
  \caption{Roughness of the fields ($1-\rho_1$) and \texttt{U-Net} MAE, in absolute value and relative to \texttt{tas}.}
  \label{tab:rough_mae}
  \centering
  \begin{tabular}{lcccc}
    \toprule
    Variable & $1-\rho_1$ & $1-\rho_1$ ($\times$\,tas) &
    MAE U-Net & MAE ($\times$\,tas) \\
    \midrule
    \texttt{tas} & 0.011 & 1.00 & 0.040 & 1.00 \\
    \texttt{u10} & 0.019 & 1.72 & 0.068 & 1.70 \\
    \texttt{v10} & 0.020 & 1.79 & 0.079 & 1.98 \\
    \texttt{pr}  & 0.045 & 4.11 & 0.163 & 4.08 \\
    \bottomrule
  \end{tabular}
\end{table}

\texttt{Precipitation} further illustrates the importance of the evaluation criterion. Its large zero-valued background makes interpolation comparatively competitive on pointwise metrics despite its poor spectral performance. Thus, a low average error does not necessarily imply that the spatial variability relevant to a downstream climate application has been faithfully reconstructed. More generally, as the spatial complexity of the variable increases, the distinction between pointwise fidelity and preservation of fine-scale variability becomes increasingly important.

\section{Discussion and Conclusion}

\paragraph{Discussion and summary.}
Evaluating climate downscaling differs from generic image super-resolution: climate fields contain information across spatial scales, whose relevance depends on the intended application. Pointwise fidelity may be essential for some applications, while others require realistic variability, extremes, or fine-scale statistics. \textit{The evaluation protocol is therefore part of the definition of the downscaling problem itself}: choosing a metric implicitly defines which notion of climate realism is rewarded. Our results illustrate this directly. \texttt{U-Net} gives the strongest pointwise and spatially aligned reconstructions, but produces smoother fields with reduced high-frequency content. \texttt{SE-OT} and \texttt{CLR} preserve substantially more spectral variability, at the cost of less accurate fine-structure localisation, while \texttt{INR-SR} is intermediate and particularly competitive for \texttt{wind} and \texttt{precipitation}. The ranking thus changes with both the metric and the variable. Beyond confirming a known trade-off, our contribution is to quantify how large it is and how systematically it shifts across metrics and variables. \textit{There is therefore no single notion of reconstruction quality for climate downscaling, and one metric is not enough.} Rather than seeking a universal ranking, a robust benchmark should make these trade-offs explicit and report complementary criteria chosen according to the intended application.

\paragraph{Limitations and future work.}
These conclusions are based on one region, one downscaling factor, and synthetically degraded low-resolution inputs. The trade-off is also partly shaped by the metrics themselves: retrieval-based methods preserve spectral content almost by construction, and none of our metrics is a dedicated displacement score, which a neighbourhood measure such as the Fractions Skill Score~\citep{roberts2008scale} would provide. Future work should test whether these trade-offs persist across regions, spatial scales, and climate-model outputs, extend the comparison to generative, diffusion-based, and physics-informed approaches, and assess whether metric-dependent rankings translate into differences in downstream climate applications.

\bibliographystyle{unsrtnat}
\bibliography{biblio}

@ARTICLE{Michelangeli2009-zb,
  title     = "Probabilistic downscaling approaches: Application to wind
               cumulative distribution functions",
  author    = "Michelangeli, P-A and Vrac, M and Loukos, H",
  journal   = "Geophys. Res. Lett.",
  publisher = "American Geophysical Union (AGU)",
  volume    =  36,
  number    =  11,
  month     =  jun,
  year      =  2009,
  language  = "en",
  url={ https://doi.org/10.1029/2009GL038401}
}

@ARTICLE{Robin2019-rz,
  title     = "Multivariate stochastic bias corrections with optimal transport",
  author    = "Robin, Yoann and Vrac, Mathieu and Naveau, Philippe and Yiou,
               Pascal",
  journal   = "Hydrol. Earth Syst. Sci.",
  publisher = "Copernicus GmbH",
  volume    =  23,
  number    =  2,
  pages     = "773--786",
  month     =  feb,
  year      =  2019,
  language  = "en"
}

@misc{dong2015imagesuperresolutionusingdeep,
      title={Image Super-Resolution Using Deep Convolutional Networks}, 
      author={Chao Dong and Chen Change Loy and Kaiming He and Xiaoou Tang},
      year={2015},
      eprint={1501.00092},
      archivePrefix={arXiv},
      primaryClass={cs.CV},
      url={https://arxiv.org/abs/1501.00092}, 
}

@misc{kim2016accurateimagesuperresolutionusing,
      title={Accurate Image Super-Resolution Using Very Deep Convolutional Networks}, 
      author={Jiwon Kim and Jung Kwon Lee and Kyoung Mu Lee},
      year={2016},
      eprint={1511.04587},
      archivePrefix={arXiv},
      primaryClass={cs.CV},
      url={https://arxiv.org/abs/1511.04587}, 
}

@misc{ronneberger2015unetconvolutionalnetworksbiomedical,
      title={U-Net: Convolutional Networks for Biomedical Image Segmentation}, 
      author={Olaf Ronneberger and Philipp Fischer and Thomas Brox},
      year={2015},
      eprint={1505.04597},
      archivePrefix={arXiv},
      primaryClass={cs.CV},
      url={https://arxiv.org/abs/1505.04597}, 
}

@misc{ledig2017photorealisticsingleimagesuperresolution,
      title={Photo-Realistic Single Image Super-Resolution Using a Generative Adversarial Network}, 
      author={Christian Ledig and Lucas Theis and Ferenc Huszar and Jose Caballero and Andrew Cunningham and Alejandro Acosta and Andrew Aitken and Alykhan Tejani and Johannes Totz and Zehan Wang and Wenzhe Shi},
      year={2017},
      eprint={1609.04802},
      archivePrefix={arXiv},
      primaryClass={cs.CV},
      url={https://arxiv.org/abs/1609.04802}, 
}

@misc{naour2024timeseriescontinuousmodeling,
      title={Time Series Continuous Modeling for Imputation and Forecasting with Implicit Neural Representations}, 
      author={Etienne Le Naour and Louis Serrano and Léon Migus and Yuan Yin and Ghislain Agoua and Nicolas Baskiotis and Patrick Gallinari and Vincent Guigue},
      year={2024},
      eprint={2306.05880},
      archivePrefix={arXiv},
      primaryClass={cs.LG},
      url={https://arxiv.org/abs/2306.05880}, 
}

@misc{sitzmann2020implicitneuralrepresentationsperiodic,
      title={Implicit Neural Representations with Periodic Activation Functions}, 
      author={Vincent Sitzmann and Julien N. P. Martel and Alexander W. Bergman and David B. Lindell and Gordon Wetzstein},
      year={2020},
      eprint={2006.09661},
      archivePrefix={arXiv},
      primaryClass={cs.CV},
      url={https://arxiv.org/abs/2006.09661}, 
}

@misc{peyré2020computationaloptimaltransport,
      title={Computational Optimal Transport}, 
      author={Gabriel Peyré and Marco Cuturi},
      year={2020},
      eprint={1803.00567},
      archivePrefix={arXiv},
      primaryClass={stat.ML},
      url={https://arxiv.org/abs/1803.00567}, 
}

@misc{cuturi2013sinkhorndistanceslightspeedcomputation,
      title={Sinkhorn Distances: Lightspeed Computation of Optimal Transportation Distances}, 
      author={Marco Cuturi},
      year={2013},
      eprint={1306.0895},
      archivePrefix={arXiv},
      primaryClass={stat.ML},
      url={https://arxiv.org/abs/1306.0895}, 
}

@misc{oord2019representationlearningcontrastivepredictive,
      title={Representation Learning with Contrastive Predictive Coding}, 
      author={Aaron van den Oord and Yazhe Li and Oriol Vinyals},
      year={2019},
      eprint={1807.03748},
      archivePrefix={arXiv},
      primaryClass={cs.LG},
      url={https://arxiv.org/abs/1807.03748}, 
}

@misc{chen2020simpleframeworkcontrastivelearning,
      title={A Simple Framework for Contrastive Learning of Visual Representations}, 
      author={Ting Chen and Simon Kornblith and Mohammad Norouzi and Geoffrey Hinton},
      year={2020},
      eprint={2002.05709},
      archivePrefix={arXiv},
      primaryClass={cs.LG},
      url={https://arxiv.org/abs/2002.05709}, 
}

@article{Zhang_2018,
   title={Road Extraction by Deep Residual U-Net},
   volume={15},
   ISSN={1558-0571},
   url={http://dx.doi.org/10.1109/LGRS.2018.2802944},
   DOI={10.1109/lgrs.2018.2802944},
   number={5},
   journal={IEEE Geoscience and Remote Sensing Letters},
   publisher={Institute of Electrical and Electronics Engineers (IEEE)},
   author={Zhang, Zhengxin and Liu, Qingjie and Wang, Yunhong},
   year={2018},
   month=May, pages={749–753} }

@article{era5,
      author = {Hersbach, Hans and Bell, Bill and Berrisford, Paul and Hirahara, Shoji and Horányi, András and Muñoz-Sabater, Joaquín and Nicolas, Julien and Peubey, Carole and Radu, Raluca and Schepers, Dinand and Simmons, Adrian and Soci, Cornel and Abdalla, Saleh and Abellan, Xavier and Balsamo, Gianpaolo and Bechtold, Peter and Biavati, Gionata and Bidlot, Jean and Bonavita, Massimo and De Chiara, Giovanna and Dahlgren, Per and Dee, Dick and Diamantakis, Michail and Dragani, Rossana and Flemming, Johannes and Forbes, Richard and Fuentes, Manuel and Geer, Alan and Haimberger, Leo and Healy, Sean and Hogan, Robin J. and Hólm, Elías and Janisková, Marta and Keeley, Sarah and Laloyaux, Patrick and Lopez, Philippe and Lupu, Cristina and Radnoti, Gabor and de Rosnay, Patricia and Rozum, Iryna and Vamborg, Freja and Villaume, Sebastien and Thépaut, Jean-Noël},
      title = {The ERA5 global reanalysis},
      journal = {Quarterly Journal of the Royal Meteorological Society},
      volume = {146},
      number = {730},
      pages = {1999-2049},
      doi = {https://doi.org/10.1002/qj.3803},
      url = {https://rmets.onlinelibrary.wiley.com/doi/abs/10.1002/qj.3803},
      eprint = {https://rmets.onlinelibrary.wiley.com/doi/pdf/10.1002/qj.3803},
      year = {2020}
      }

@article{era5_1940,
      author = {Soci, Cornel and Hersbach, Hans and Simmons, Adrian and Poli, Paul and Bell, Bill and Berrisford, Paul and Horányi, András and Muñoz-Sabater, Joaquín and Nicolas, Julien and Radu, Raluca and Schepers, Dinand and Villaume, Sebastien and Haimberger, Leopold and Woollen, Jack and Buontempo, Carlo and Thépaut, Jean-Noël},
      title = {The ERA5 global reanalysis from 1940 to 2022},
      journal = {Quarterly Journal of the Royal Meteorological Society},
      volume = {150},
      number = {764},
      pages = {4014-4048},
      doi = {https://doi.org/10.1002/qj.4803},
      url = {https://rmets.onlinelibrary.wiley.com/doi/abs/10.1002/qj.4803},
      eprint = {https://rmets.onlinelibrary.wiley.com/doi/pdf/10.1002/qj.4803},
      year = {2024}
      }

@Article{Willmott2005,
      author={Willmott, C. J.
      and Matsuura, K.},
      title={Advantages of the mean absolute error (MAE) over the root mean square error (RMSE) in assessing average model performance},
      journal={Climate Research},
      year={2005},
      volume={30},
      pages={79-82},
      doi={10.3354/cr030079},
      url={https://www.int-res.com/journals/cr/articles/cr030079},
      url={https://doi.org/10.3354/cr030079}
      }

@ARTICLE{ssim,
      author={Zhou Wang and Bovik, A.C. and Sheikh, H.R. and Simoncelli, E.P.},
      journal={IEEE Transactions on Image Processing}, 
      title={Image quality assessment: from error visibility to structural similarity}, 
      year={2004},
      volume={13},
      number={4},
      pages={600-612},
      doi={10.1109/TIP.2003.819861}}

@inbook{optimtransport,
      author = {Villani, Cédric},
      year = {2008},
      month = {01},
      pages = {xxii+973},
      title = {Optimal transport -- Old and new},
      volume = {338},
      doi = {10.1007/978-3-540-71050-9}
      }

@misc{arjovsky2017wassersteingan,
      title={Wasserstein GAN}, 
      author={Martin Arjovsky and Soumith Chintala and Léon Bottou},
      year={2017},
      eprint={1701.07875},
      archivePrefix={arXiv},
      primaryClass={stat.ML},
      url={https://arxiv.org/abs/1701.07875}, 
}

@article{Harris_2022,
      title={A Generative Deep Learning Approach to Stochastic Downscaling of Precipitation Forecasts},
      volume={14},
      ISSN={1942-2466},
      url={http://dx.doi.org/10.1029/2022MS003120},
      DOI={10.1029/2022ms003120},
      number={10},
      journal={Journal of Advances in Modeling Earth Systems},
      publisher={American Geophysical Union (AGU)},
      author={Harris, Lucy and McRae, Andrew T. T. and Chantry, Matthew and Dueben, Peter D. and Palmer, Tim N.},
      year={2022},
      month=Oct }

@ARTICLE{gradient,
  author={Xue, Wufeng and Zhang, Lei and Mou, Xuanqin and Bovik, Alan C.},
  journal={IEEE Transactions on Image Processing}, 
  title={Gradient Magnitude Similarity Deviation: A Highly Efficient Perceptual Image Quality Index}, 
  year={2014},
  volume={23},
  number={2},
  pages={684-695},
  doi={10.1109/TIP.2013.2293423}}

@article{wilby2013statistical,
  title={The Statistical Downscaling Model: insights from one decade of application.},
  author={Wilby, Robert L and Dawson, Christian W},
  journal={International Journal of Climatology},
  volume={33},
  number={7},
  pages={1707},
  year={2013}
}

@article{xu2019dynamical,
  title={Dynamical downscaling of regional climate: A review of methods and limitations},
  author={Xu, Zhongfeng and Han, Ying and Yang, Zongliang},
  journal={Science China Earth Sciences},
  volume={62},
  number={2},
  pages={365--375},
  year={2019},
  publisher={Springer}
}

@article{hong2014dynamical,
  title={Dynamical downscaling: Fundamental issues from an NWP point of view and recommendations},
  author={Hong, Song-You and Kanamitsu, Masao},
  journal={Asia-Pacific Journal of Atmospheric Sciences},
  volume={50},
  number={1},
  pages={83--104},
  year={2014},
  publisher={Springer}
}

@book{benestad2008empirical,
  title={Empirical-statistical downscaling},
  author={Benestad, Rasmus E and Chen, Deliang and Hanssen-Bauer, Inger},
  year={2008},
  publisher={World Scientific Publishing Company}
}

@inproceedings{blau2018perception,
  title={The perception-distortion tradeoff},
  author={Blau, Yochai and Michaeli, Tomer},
  booktitle={2018 IEEE/CVF Conference on Computer Vision and Pattern Recognition},
  pages={6228--6237},
  year={2018},
  organization={IEEE}
}

@article{freirich2021theory,
  title={A theory of the distortion-perception tradeoff in wasserstein space},
  author={Freirich, Dror and Michaeli, Tomer and Meir, Ron},
  journal={Advances in Neural Information Processing Systems},
  volume={34},
  pages={25661--25672},
  year={2021}
}

@article{roberts2008scale,
  title={Scale-selective verification of rainfall accumulations from high-resolution forecasts of convective events},
  author={Roberts, Nigel M and Lean, Humphrey W},
  journal={Monthly Weather Review},
  volume={136},
  number={1},
  pages={78--97},
  year={2008}
}

@article{zeraatkar2026frequency,
  title={Frequency-aware vision transformers for high-fidelity super-resolution of Earth system models},
  author={Zeraatkar, Ehsan and Faroughi, Salah A and Te{\v{s}}i{\'c}, Jelena},
  journal={Scientific Reports},
  volume={16},
  number={1},
  pages={10363},
  year={2026},
  publisher={Nature Publishing Group UK London}
}

@inproceedings{NEURIPS2023_df543023,
 author = {Serrano, Louis and Le Boudec, Lise and Kassa\"{\i} Koupa\"{\i}, Armand and Wang, Thomas X and Yin, Yuan and Vittaut, Jean-No\"{e}l and Gallinari, Patrick},
 booktitle = {Advances in Neural Information Processing Systems},
 doi = {10.52202/075280-3093},
 editor = {A. Oh and T. Naumann and A. Globerson and K. Saenko and M. Hardt and S. Levine},
 pages = {70581--70611},
 publisher = {Curran Associates, Inc.},
 title = {Operator Learning with Neural Fields: Tackling PDEs on General Geometries},
 url = {https://proceedings.neurips.cc/paper_files/paper/2023/file/df54302388bbc145aacaa1a54a4a5933-Paper-Conference.pdf},
 volume = {36},
 year = {2023}
}

\clearpage
\appendix
\section{Appendix}

In the remainder of this document, we provide a description of the problem to be solved, 
a detailed description of the super-resolution methods as well as of the metrics used 
to test these methods. Finally, we conclude with the whole set of numerical results obtained
as well as some qualitative examples of the variables used. 

\subsection{Formalisation of the problem}

Statistical downscaling is treated here as an image super-resolution problem. Each
climate data is a multivariate field with $C$ channels (combination of the climate variables 
\texttt{tas}, \texttt{u10}, \texttt{v10}, \texttt{pr}) defined
on a regular spatial grid. We denote $x_{low} \in \mathbb{R}^{C \times H_{low} \times W_{low}}$ a
low-resolution (LR) image on the coarse grid $\mathcal{G}_{low}$, and
$x_{high} \in \mathbb{R}^{C \times H_{high} \times W_{high}}$ the corresponding high-resolution (HR) image on
the fine grid $\mathcal{G}_{high}$, with $H_{high} \gg H_{low}$ and $W_{high} \gg W_{low}$.

The objective is to build a mapping which, from an LR image, produces an estimate
$\hat{x}_{high}$ of the HR image, as close as possible to the ground truth $x_{high}$. The 
reconstruction quality is measured by a loss function $\mathcal{L}(\hat{x}_{high}, x_{high})$.

All the methods rely on a training/reference set of $N$ paired LR/HR couples,
$\{(x_{low}^{(j)}, x_{high}^{(j)})\}_{j=1}^{N}$, where the superscript $(j)$ indexes a reference example. At
inference, we denote $x_{low}^{(i)}$ a query image and $\hat{x}_{high}^{(i)}$ its reconstruction. The
parametric methods (residual U-Net, INR-SR) are trained by batches $\mathcal{B}$ drawn from this set, whereas the
nearest-neighbour methods (sparse optimal transport, contrastive latent retrieval) use
directly the reference couples as a dictionary. These notations are common to all the
subsections that follow.

\subsection{Methods}\label{app:methods}

\subsubsection{Residual U-Net method for super-resolution}\label{app:unet}

\begin{wrapfigure}[28]{r}{0.55\linewidth}
  \centering
  \resizebox{\linewidth}{!}{%
\begin{tikzpicture}[
        node distance=0.8cm and 1cm,
        block/.style={
            rectangle,
            draw,
            thick,
            fill=blue!20,
            text width=7em,
            align=center,
            minimum height=2.5em,
            rounded corners,
        },
        bottleneck/.style={
            rectangle,
            draw,
            thick,
            fill=green!30,
            text width=8em,
            align=center,
            minimum height=2.5em,
            rounded corners,
        },
        io/.style={
            trapezium,
            trapezium left angle=70,
            trapezium right angle=110,
            draw,
            thick,
            fill=gray!20,
            align=center,
            minimum height=2em,
        },
        op/.style={
            rectangle,
            draw,
            thick,
            fill=orange!30,
            align=center,
            minimum height=1.5em
        },
        add/.style={
            circle,
            draw,
            thick,
            fill=yellow!40,
            minimum size=1.4em,
            inner sep=0pt,
            font=\large
        },
        arrow/.style={-latex, thick},
        down_arrow/.style={-latex, thick, red!80!black},
        up_arrow/.style={-latex, thick, green!60!black},
        skip_arrow/.style={-latex, thick, gray, dashed},
        residual_arrow/.style={-latex, very thick, violet}
    ]


    \node[bottleneck] (bneck) {\textbf{Bottleneck} \\ \texttt{512 channels}};

    \node[block, above left=1cm and -0.5cm of bneck] (enc2) {\textbf{Encoder 3}\\ \texttt{256 channels}};
    \node[block, above right=1cm and -0.5cm of bneck] (dec2) {\textbf{Decoder 1}\\ \texttt{256 channels}};

    \node[block, above=of enc2] (enc1) {\textbf{Encoder 2}\\ \texttt{128 channels}};
    \node[block, above=of dec2] (dec3) {\textbf{Decoder 2}\\ \texttt{128 channels}};

    \node[block, above=of enc1] (in_conv) {\textbf{Encoder 1}\\ \texttt{64 channels}};
    \node[block, above=of dec3] (dec4) {\textbf{Decoder 3}\\ \texttt{64 channels}};

    \node[op, above=of in_conv] (conv_in) {\texttt{Conv2d}};
    \node[op, above=of conv_in] (upsize) {\texttt{Upsize}};
    \node[io, above=0.4cm of upsize] (input) {Input\\ \tiny ($B, C, H_{low}, W_{low}$)};

    \node[op, above=of dec4] (out_conv) {\texttt{Conv2d}};
    \node[add, at={(out_conv |- upsize)}] (add) {$+$};
    \node[io, at={(add |- input)}] (output) {Output\\ \tiny ($B, C, H_{high}, W_{high}$)};

    \draw[arrow] (input) -- node[right] {\tiny $C, H_{low}, W_{low}$} (upsize);
    \draw[arrow] (upsize) -- node[right] {\tiny $C, H_{2^n}, W_{2^n}$} (conv_in);
    \draw[arrow] (conv_in) -- node[right] {\tiny $64, H_{2^n}, W_{2^n}$} (in_conv);
    \draw[down_arrow] (in_conv) -- node[right] {\tiny $128, H_{2^n}/2, W_{2^n}/2$} (enc1);
    \draw[down_arrow] (enc1) -- node[right] {\tiny $256, H_{2^n}/4, W_{2^n}/4$} (enc2);
    \draw[down_arrow] (enc2) -- node[left] {\tiny $512, H_{2^n}/8, W_{2^n}/8$} (bneck);

    \draw[up_arrow] (bneck) -- node[right] {\tiny 512, $H_{2^n}/8, W_{2^n}/8$} (dec2);
    \draw[up_arrow] (dec2) -- node[left] {\tiny 256, $H_{2^n}/4, W_{2^n}/4$} (dec3);
    \draw[up_arrow] (dec3) -- node[left] {\tiny 128, $H_{2^n}/2, W_{2^n}/2$} (dec4);
    \draw[arrow] (dec4) -- node[left] {\tiny $64, H_{2^n}, W_{2^n}$} (out_conv);
    \draw[arrow] (out_conv) -- node[left] {\tiny $C, H_{2^n}, W_{2^n}$} (add);
    \draw[arrow] (add) -- node[left] {\tiny $C, H_{high}, W_{high}$} (output);

    \draw[skip_arrow] (enc2.east) -- node[above, midway] {Skip} (dec2.west);
    \draw[skip_arrow] (enc1.east) -- node[above, midway] {Skip} (dec3.west);
    \draw[skip_arrow] (in_conv.east) -- node[above, midway] {Skip} (dec4.west);

    \draw[residual_arrow] (upsize.east)
        -- node[above, midway] {Residual}
           node[below, midway] {\tiny $C, H_{2^n}, W_{2^n}$} (add.west);

    \end{tikzpicture}%
  }
\end{wrapfigure}
The guiding principle of this method is \emph{residual learning}. 
The low-resolution input image is brought to the dimension of the target grid by
a \emph{bicubic} interpolation, which provides a good-quality base image. A U-Net network then predicts
the residual, that is, the high-frequency correction to add to this base. The final output is the
sum of the base interpolation and of the predicted residual. This decomposition stabilises the training, the network
focusing on the ``missing information'' instead of recreating the whole image.

The encoder comprises three contraction levels. This deliberately limited depth is dictated by the
very low resolution of the inputs (of the order of $9\times7$): each level halves the spatial
resolution, i.e. a division by $8$ at the bottleneck, which preserves a feature
map still spatially coherent — a deeper U-Net would reduce the image below the pixel. The
downsampling is performed by strided convolutions (rather than \textit{max-pooling}), so that
the network itself learns the way to compress the information. Robustness to arbitrary dimensions
(not multiples of $2$) is ensured by two mechanisms: a reflection padding
for the border convolutions, and a \emph{dynamic} upsampling in the decoder, where each feature
map is resized exactly to the size of the skip connection
to which it is concatenated.

    Let $\mathrm{Up}^{\mathrm{bic}}_{\mathcal{G}_{high}}$ be the bicubic interpolation towards the target grid. 
    Let $U_\theta$ be the residual U-Net network, of parameters $\theta$, which predicts the high-frequency
    residual. We have:
    
    \begin{equation}
        b = \mathrm{Up}^{\mathrm{bic}}_{\mathcal{G}_{high}}(x_{low})
        \qquad \text{and} \qquad
        \hat{x}_{high} = b + U_\theta(b)
        \label{eq:unet_residual}
    \end{equation}

    The network is trained to minimise the mean absolute error ($\ell_1$ loss) between the prediction and the
    ground truth over a batch $\mathcal{B}$,
    \label{eq:unet_loss}
    \begin{equation}
        \mathcal{L}(\theta) = \frac{1}{|\mathcal{B}|} \sum_{j \in \mathcal{B}}
        \big\lVert \hat{x}_{high}^{(j)} - x_{high}^{(j)} \big\rVert_1 
    \end{equation}
    which amounts to making the network learn the true residual
    $x_{high}^{(j)} - b^{(j)}$. The optimisation is carried out by the AdamW algorithm. The training
    and inference procedures are described by algorithms~\ref{alg:train_unet} and~\ref{alg:inf_unet}.

    \vspace{1em}
    \begin{algorithm}[H]
        \label{alg:train_unet}
        \SetAlgoLined
        \KwData{Training data ${(x_{low}^{(j)}, x_{high}^{(j)})}$}
        \KwResult{Trained parameters $\theta$ of the residual U-Net}
        \While{not converged}{
            Take a batch $\mathcal{B}$ of data ${(x_{low}^{(j)}, x_{high}^{(j)})}_{j \in \mathcal{B}}$\;
            Base interpolation $b^{(j)} \leftarrow \mathrm{Up}^{\mathrm{bic}}_{\mathcal{G}_{high}}({x_{low}^{(j)}})$\;
            Prediction $\hat{x}_{high}^{(j)} \leftarrow b^{(j)} + U_\theta(b^{(j)})$\;
            Loss $\mathcal{L}(\theta) \leftarrow \frac{1}{|\mathcal{B}|} \sum_{j \in \mathcal{B}} \lVert \hat{x}_{high}^{(j)} - {x_{high}^{(j)}} \rVert_1$\;
            Update $\theta \leftarrow \theta - \eta\, \nabla_{\theta}\, \mathcal{L}(\theta)$ \quad (AdamW)\;
        }
        \caption{Training of the residual U-Net}
    \end{algorithm}
    \vspace{1em}

    \begin{algorithm}[H]
        \label{alg:inf_unet}
        \SetAlgoLined
        \KwData{Trained network $U_\theta$, query image ${x_{low}^{(i)}}$, target grid $\mathcal{G}_{high}$}
        \KwResult{Reconstructed high-resolution image $\hat{x}_{high}^{(i)}$}
        Base interpolation $b^{(i)} \leftarrow \mathrm{Up}^{\mathrm{bic}}_{\mathcal{G}_{high}}({x_{low}^{(i)}})$\;
        Residual reconstruction $\hat{x}_{high}^{(i)} \leftarrow b^{(i)} + U_\theta(b^{(i)})$\;
        \caption{Inference of the residual U-Net}
    \end{algorithm}
    \vspace{1em}

\subsubsection{INR Super Resolution Method}\label{app:INR-SR}

    INR Super Resolution (INR-SR) is an evolution of TimeFlow\citep{naour2024timeseriescontinuousmodeling}.
    TimeFlow relies on an implicit representation network (INR) whose behaviour is driven by a
    latent code: at inference, the code inferred from an LR image makes it possible to generate the image on
    the target HR grid.

    During training, INR-SR uses an LR and HR image pair at the same time. 
    The latent space learns to 
    move the INR ``to the right place'' according to the low-resolution images, 
    that is, in such a way that the INR can respond with respect to the 
    high-resolution climate image. This is the object of the two following algorithms \ref{alg:train_timeflow_diff} and 
    \ref{alg:inf_timeflow_diff}.

    More precisely, 
    in the inner loop, the algorithm carries out the convergence of the latent code $z^{(j)}$ 
    using the low-resolution 
    climate image $x_{low}^{(j)}$ and therefore a low-resolution grid $\mathcal{G}_{low}^{(j)}$. One seeks to minimise the loss 
    function $\mathcal{L}$ between the initial image $x_{low}^{(j)}$ and the image computed by the INR, given that the weights 
    of the modulation networks and the INR are frozen at that precise moment. 
    Once the latent codes $z^{(j)}$ obtained and fixed, one can then update the model weights (INR and the modulation) 
    by comparing the true high-resolution climate image $x_{high}^{(j)}$ with the computed image $\hat{x}_{high}^{(i)}$. 
    The modifications
    with respect to the initial algorithm\citep{naour2024timeseriescontinuousmodeling} are shown in pink.

    \vspace{1em}
    \begin{algorithm}[H]
        \label{alg:train_timeflow_diff}
        \SetAlgoLined 
        \KwData{Training data $\textcolor{magenta}{(x_{low}^{(j)}, x_{high}^{(j)})}$}  
        \KwResult{Trained parameters $\theta, w$}  
        \While{not converged}{  
            Take a batch $\mathcal{B}$ of data $\textcolor{magenta}{(x_{low}^{(j)}, x_{high}^{(j)})}_{j \in \mathcal{B}}$\;  
            Initialise the latent codes to zero $z^{(j)} \leftarrow 0, \forall j \in \mathcal{B}$\;  

            // inner loop for encoding

            \For{$j \in \mathcal{B}$ and step $k \in \{1,\ldots,K\}$}{  
                Update the code $z^{(j)} \leftarrow z^{(j)} - \alpha \nabla_{z^{(j)}} \mathcal{L}_{\textcolor{magenta}{\mathcal{G}_{low}^{(j)}}}(f_{\theta,h_w(z^{(j)})}, \textcolor{magenta}{x_{low}^{(j)}})$\;  
            }

            // outer loop step

            $[\theta, w] \leftarrow [\theta, w] - \eta \nabla_{[\theta,w]} \frac{1}{|\mathcal{B}|} \sum_{j \in \mathcal{B}} \mathcal{L}_{\textcolor{magenta}{\mathcal{G}_{high}^{(j)}}}(f_{\theta,h_w(z^{(j)})}, \textcolor{magenta}{x_{high}^{(j)}})$\;  
        }
        \caption{Training of INR Super Resolution}
    \end{algorithm}
    \vspace{1em}

    For inference, we use the low-resolution climate image $x_{low}^{(j)}$ to obtain 
    the corresponding latent code $z^{*(j)}$. It is then possible to obtain any climate image following a chosen 
    grid, for example the grid $\mathcal{G}_{high}$ to come back to a climate image equivalent to the images of the 
    training phase, but any other configuration in the general space $\mathcal{G}^{*(j)}$ is possible.

    \vspace{1em}
    \begin{algorithm}[H]
        \label{alg:inf_timeflow_diff}
        \SetAlgoLined 
        
        For the $j^{\text{th}}$ series  $\textcolor{magenta}{(x_{low}^{(j)})}$, initialise the latent codes to zero $z^{*(j)} \leftarrow 0$\;
        \For{step $\in \{1,\ldots,K\}$}{
            $z^{*(j)} \leftarrow z^{*(j)} - \alpha \nabla_{z^{*(j)}} \mathcal{L}_{\textcolor{magenta}{\mathcal{G}_{low}^{(j)}}}(f_{\theta,h_w(z^{*(j)})}, \textcolor{magenta}{x_{low}^{(j)}})$\;
        }
        Compute $f_{\theta,h_w(z^{*(j)})}(g_x, g_y)$ for every couple $(g_x, g_y) \in \mathcal{G}_{high}^{(j)} ( \mathcal{G}^{*(j)})$\;

        \caption{Inference of INR Super Resolution}
    \end{algorithm}
    \vspace{1em}    

\subsubsection{Sparse Entropic Optimal Transport Method}\label{app:SE-OT}

    This method is \emph{non-parametric}, it does not involve a training phase and uses
    directly the reference set as a dictionary of anchors. The objective is to express the query
    $x_{low}^{(i)}$ as an optimal combination of the LR anchors, then to transport this structure towards
    the HR space in order to reconstruct $\hat{x}_{high}^{(i)}$.

    \textbf{Regularised optimal transport}: 
    the cost matrix is defined by the squared Euclidean distance between the query and each anchor
    in the low-resolution space:
    \begin{equation}
        C_{ij} = \left\lVert x_{low}^{(i)} - x_{low}^{(j)} \right\rVert_2^2 
    \end{equation}
    The transport plan $P$ is obtained by minimising the total transport cost,
    expressed by the Frobenius product $\langle P, C\rangle_F = \sum_{i,j} P_{ij} C_{ij}$, regularised by
    the Shannon entropy $H(P) = \sum_{i,j} P_{ij}\,(\log P_{ij} - 1)$~\citep{cuturi2013sinkhorndistanceslightspeedcomputation, peyré2020computationaloptimaltransport}, i.e.:
    \begin{equation}
        \min_{P \in U(a,b)} \; \langle P, C\rangle_F + \epsilon\, H(P) 
    \end{equation}
    where $\epsilon$ is the smoothing temperature and $U(a,b)$ the set of plans respecting the
    mass-conservation constraints $\sum_j P_{ij} = a_i$ and $\sum_i P_{ij} = b_j$.

    \textbf{Gibbs kernel}:
    this is a problem of minimising a function under constraints. The resolution requires
    the introduction of the Lagrange multipliers $\alpha_i$ and $\beta_j$ associated with these constraints, then
    the cancellation of the first derivative of the Lagrangian $\partial \mathcal{L} / \partial P_{ij} = C_{ij} +
    \epsilon \log P_{ij} - \alpha_i - \beta_j = 0$, lead to the fundamental form of the optimal plan
    $P_{ij} = \exp\!\big((\alpha_i + \beta_j - C_{ij})/\epsilon\big)$. By factorising, the
    \emph{Gibbs kernel} $K_{ij} = \exp(-C_{ij}/\epsilon)$ emerges, which converts the distances into affinity
    measures, and one obtains the solution:
    \begin{equation}
        P = \operatorname{diag}(u)\, K\, \operatorname{diag}(v), \qquad
        u_i = \exp(\alpha_i/\epsilon), \quad v_j = \exp(\beta_j/\epsilon)
    \end{equation}

    \textbf{Sparse barycentric mapping}: we want to project our input onto the reference basis.
    Instead of iterating to find u and v (Sinkhorn algorithm), the mass-conservation constraint 
    for the columns ($v_j = 1$) is dropped and one normalises by row ($\sum_j P_{ij} = 1$).
    Which amounts to introducing a softmax:
    \begin{equation}
        w_{ij} = \frac{\exp(-C_{ij}/\epsilon)}{\sum_{k} \exp(-C_{ik}/\epsilon)}
        \label{eq:ot_softmax}
    \end{equation}
    To preserve the sharpness of the reconstruction and avoid the mixing of semantically distant examples
    (amounting to a regression towards the mean), the transport is made \emph{sparse}. The support
    of $w_{i\cdot}$ is restricted to the $k$ nearest neighbours $\mathcal{N}_k(i)$ of the query, by setting $C_{ij} = +\infty$
    (hence $K_{ij} = 0$) for $j \notin \mathcal{N}_k(i)$.

    \textbf{Reconstruction}: 
    under the assumption of a local isometry between the low- and high-resolution manifolds, the
    transport structure computed on $\mathcal{G}_{low}$ is reused on $\mathcal{G}_{high}$. The high-resolution
    image is the barycentre of the HR anchors weighted by the transport plan:
    \begin{equation}
        \hat{x}_{high}^{(i)} = \sum_{j \in \mathcal{N}_k(i)} w_{ij}\, x_{high}^{(j)}
        \label{eq:ot_bary}
    \end{equation}
    The parameter $\epsilon$ controls the curvature of the Gibbs distribution (a small $\epsilon$ finds a
    solution on the nearest neighbour, a large $\epsilon$ diffuses it), while $k$ fixes the extent of the
    neighbourhood. The Gibbs kernel being very sensitive to the scale of the distances, the costs are normalised
    by their median before the softmax for numerical stability. Algorithm~\ref{alg:inf_ot_sr} gives the inference procedure.

    \vspace{1em}
    \begin{algorithm}[H]
        \label{alg:inf_ot_sr}
        \SetAlgoLined
        \KwData{Reference data ${(x_{low}^{(j)}, x_{high}^{(j)})}$, query image ${x_{low}^{(i)}}$, regularisation $\epsilon$, neighbourhood size $k$}
        \KwResult{Reconstructed high-resolution image $\hat{x}_{high}^{(i)}$}
        \For{$j \in \{1,\ldots,N\}$}{
            Transport cost $C_{ij} \leftarrow \left\lVert {x_{low}^{(i)}} - {x_{low}^{(j)}} \right\rVert_2^2$\;
        }
        Stabilisation: $C_{ij} \leftarrow C_{ij} / \big(\operatorname{med}_j C_{ij} + \delta\big)$\;
        Top-k selection: $\mathcal{N}_k(i) \leftarrow$ indices of the $k$ smallest costs $C_{ij}$\;
        Sparse masking: $C_{ij} \leftarrow +\infty$ for $j \notin \mathcal{N}_k(i)$\;
        Barycentric weights: $w_{ij} \leftarrow \dfrac{\exp(-C_{ij}/\epsilon)}{\sum_{k} \exp(-C_{ik}/\epsilon)}$\;
        HR reconstruction: $\hat{x}_{high}^{(i)} \leftarrow \sum_{j \in \mathcal{N}_k(i)} w_{ij}\, {x_{high}^{(j)}}$\;
        \caption{Inference of Sparse Entropic Optimal Transport}
    \end{algorithm}
    \vspace{1em}

\subsubsection{Contrastive Latent Retrieval Method}\label{app:clr}

    This method is \emph{hybrid}. A parametric phase learns a representation space of the LR
    images, followed by a non-parametric reconstruction phase by retrieval of the nearest neighbours in
    this space, the reference set serving as a dictionary.

    \paragraph{Latent space and cosine similarity.}
    A convolutional encoder $E_\phi$, of parameters $\phi$, projects a low-resolution image onto a normalised
    latent vector,
    \begin{equation}
        z^{(j)} = \frac{E_\phi(x_{low}^{(j)})}{\lVert E_\phi(x_{low}^{(j)}) \rVert_2} \in \mathbb{R}^{d}
        \qquad \lVert z^{(j)} \rVert_2 = 1 
    \end{equation}
    so that the proximity between two images is measured by the cosine similarity $s_{ij} = \langle z^{(i)},
    z^{(j)} \rangle$.

    \paragraph{Contrastive learning.}
    The encoder $E_\phi$ is trained in a self-supervised way. Each image $x_{low}^{(j)}$ is paired with a version
    augmented by Gaussian noise $\tilde{x}_{low}^{(j)} = x_{low}^{(j)} + \sigma\,\eta$, $\eta \sim \mathcal{N}(0, I)$,
    whose normalised latent code is denoted $\tilde{z}^{(j)}$. Over a batch $\mathcal{B}$, the InfoNCE-type
    contrastive loss~\citep{oord2019representationlearningcontrastivepredictive} brings each image closer to its own augmentation
    while pushing it away from the other examples of the batch,
    \begin{equation}
        \mathcal{L}_{\mathrm{NCE}} = -\frac{1}{|\mathcal{B}|} \sum_{j \in \mathcal{B}}
        \log \frac{\exp\!\big(\langle z^{(j)}, \tilde{z}^{(j)} \rangle / \tau\big)}
                  {\sum_{k \in \mathcal{B}} \exp\!\big(\langle z^{(j)}, \tilde{z}^{(k)} \rangle / \tau\big)}
        \label{eq:infonce}
    \end{equation}
    where $\tau$ is the contrastive temperature. This objective structures the latent space so that semantically
    close low-resolution climate images are projected onto neighbouring points.

    \paragraph{Indexing and reconstruction.}
    Once $\phi$ frozen, the whole reference set is indexed into a memory of couples
    $\{(z^{(j)}, x_{high}^{(j)})\}_{j=1}^{N}$. At inference, the query $x_{low}^{(i)}$ is encoded into $z^{(i)}$, then one
    selects the set of the $k$ neighbours of highest cosine similarity. These similarities
    are converted into interpolation weights by a softmax of temperature $T$,
    \begin{equation}
        w_{ij} = \frac{\exp(s_{ij}/T)}{\sum_{p \in \mathcal{N}_k(i)} \exp(s_{ip}/T)} 
        \qquad j \in \mathcal{N}_k(i) 
        \label{eq:knn_softmax}
    \end{equation}
    and the high-resolution image is reconstructed as the barycentre of the corresponding HR anchors,
    \begin{equation}
        \hat{x}_{high}^{(i)} = \sum_{j \in \mathcal{N}_k(i)} w_{ij}\, x_{high}^{(j)} 
        \label{eq:knn_bary}
    \end{equation}
    The latent dimension $d$ controls the discrimination power of the space, a dimension that must be limited at the risk
    of diluting the data in a too large space.  
    The number of neighbours $k$ arbitrates between fineness ($k$ small) and robustness ($k$ large).
    The temperature $T$ tunes the reconstruction of the HR datum, from the nearest neighbour ($T \to 0$) towards a 
    uniform distribution ($T$ large). The two phases are detailed by
    algorithms~\ref{alg:train_knn} and~\ref{alg:inf_knn}.

    \vspace{1em}
    \begin{algorithm}[H]
        \label{alg:train_knn}
        \SetAlgoLined
        \KwData{Training low-resolution images ${x_{low}^{(j)}}$}
        \KwResult{Trained parameters of the encoder $\phi$}
        \While{not converged}{
            Take a batch $\mathcal{B}$ of images ${x_{low}^{(j)}}_{j \in \mathcal{B}}$\;
            Noised augmentation $\tilde{x}_{low}^{(j)} \leftarrow {x_{low}^{(j)}} + \sigma\,\eta^{(j)}, \; \eta^{(j)} \sim \mathcal{N}(0, I)$\;
            Normalised encoding $z^{(j)} \leftarrow E_\phi({x_{low}^{(j)}}) / \lVert \cdot \rVert_2$ and $\tilde{z}^{(j)} \leftarrow E_\phi(\tilde{x}_{low}^{(j)}) / \lVert \cdot \rVert_2$\;
            Update $\phi \leftarrow \phi - \eta \nabla_{\phi}\, \mathcal{L}_{\mathrm{NCE}}\big(\{z^{(j)}, \tilde{z}^{(j)}\}_{j \in \mathcal{B}}\big)$\;
        }
        \caption{Contrastive training of the latent encoder}
    \end{algorithm}

    \vspace{1em} 
    \begin{algorithm}[H]
        \label{alg:inf_knn}
        \SetAlgoLined
        \KwData{Encoder $E_\phi$, reference data ${(x_{low}^{(j)}, x_{high}^{(j)})}$, query ${x_{low}^{(i)}}$, number of neighbours $k$, temperature $T$}
        \KwResult{Reconstructed high-resolution image $\hat{x}_{high}^{(i)}$}
        \For{$j \in \{1,\ldots,N\}$}{
            Indexing $z^{(j)} \leftarrow E_\phi({x_{low}^{(j)}}) / \lVert \cdot \rVert_2$\;
        }
        Query encoding $z^{(i)} \leftarrow E_\phi({x_{low}^{(i)}}) / \lVert \cdot \rVert_2$\;
        Cosine similarities $s_{ij} \leftarrow \langle z^{(i)}, z^{(j)} \rangle$ for all $j$\;
        Top-$k$ selection: $\mathcal{N}_k(i) \leftarrow$ indices of the $k$ largest similarities $s_{ij}$\;
        Interpolation weights: $w_{ij} \leftarrow \dfrac{\exp(s_{ij}/T)}{\sum_{p \in \mathcal{N}_k(i)} \exp(s_{ip}/T)}$\;
        HR reconstruction: $\hat{x}_{high}^{(i)} \leftarrow \sum_{j \in \mathcal{N}_k(i)} w_{ij}\, {x_{high}^{(j)}}$\;
        \caption{Inference by contrastive latent retrieval}
    \end{algorithm}
    \vspace{1em}

\subsection{Metrics}\label{app:metrics}

The five retained metrics evaluate complementary aspects of the reconstruction $\hat{x}_{high}$ with respect
to the ground truth $x_{high}$, such as pointwise fidelity,
local structure, value distribution, spectral content and edge sharpness. In all the
following, we denote $M = C.\,H_{high}.\,W_{high}$ the total number of values (the channels multiplied by the number of
pixels), $p$ a value index and $\hat{x}_{high}(p)$, $x_{high}(p)$ the corresponding values. Each score
compares a reconstruction to its truth, then is averaged over the test set. A
lower score reflects a better reconstruction ($\downarrow$), except for the SSIM which is maximised ($\uparrow$). For
each of them, we indicate the criteria that differentiate them and that justify using them jointly.

\paragraph{Mean absolute error (MAE, $\downarrow$):}\label{app:mae}
the MAE is the mean of the absolute deviations between prediction and truth:
\begin{equation}
\mathrm{MAE}(\hat{x}_{high}, x_{high}) = \frac{1}{M} \sum_{p} \big\lvert \hat{x}_{high}(p) - x_{high}(p) \big\rvert 
\end{equation}
By confronting each pixel with its counterpart, it offers a direct and easily interpretable measure of the
overall fidelity, but it remains indifferent to the way the error is distributed in space. On the
frequency plane, it is essentially governed by the low frequencies. It is the large structures and a
possible mean bias that weigh the most in the sum. It is on the other hand poorly discriminative with respect to the
high frequencies, because, under uncertainty, the field that minimises the $\ell_1$ error is a smoothed field. A
blurry reconstruction, impoverished in fine details, can therefore keep a low MAE.

\paragraph{Structural Similarity Index Measure (SSIM, $\uparrow$):}\label{app:ssim}
the SSIM aims to reflect the similarity as a human observer would perceive it. It is evaluated over
sliding windows $w$ running through the image and compares, for each window, the luminance, the contrast and the structure 
for the prediction
$\hat{x}_w$ and the truth $x_w$:
\begin{equation*}
\mathrm{SSIM}(\hat{x}_w, x_w) = l(\hat{x}_w, x_w)\cdot c(\hat{x}_w, x_w)\cdot s(\hat{x}_w, x_w)
\end{equation*}
where the luminance compares the means, the contrast the standard deviations and the structure the correlation:
\begin{align*}
l(\hat{x}_w, x_w) &= \frac{2\mu_{\hat{x}_w}\mu_{x_w} + C_1}{\mu_{\hat{x}_w}^2 + \mu_{x_w}^2 + C_1} \\
c(\hat{x}_w, x_w) &= \frac{2\sigma_{\hat{x}_w}\sigma_{x_w} + C_2}{\sigma_{\hat{x}_w}^2 + \sigma_{x_w}^2 + C_2} \\
s(\hat{x}_w, x_w) &= \frac{\sigma_{\hat{x}_w x_w} + C_3}{\sigma_{\hat{x}_w}\sigma_{x_w} + C_3}
\end{align*}
with $\mu$ the local means, $\sigma$ the standard deviations, $\sigma_{\hat{x}_w x_w}$ the covariance, and
$C_1, C_2, C_3$ small constants avoiding divisions by zero. The global SSIM is the mean of the window
scores, $\tfrac{1}{|\mathcal{W}|}\sum_{w} \mathrm{SSIM}(\hat{x}_w, x_w)$. In practice, one sets
$C_3 = C_2/2$:
\begin{equation}
\mathrm{SSIM} = \frac{1}{|\mathcal{W}|} \sum_{w \in \mathcal{W}}
\frac{(2\mu_{\hat{x}_w}\mu_{x_w} + C_1)\,(2\sigma_{\hat{x}_w x_w} + C_2)}
     {(\mu_{\hat{x}_w}^2 + \mu_{x_w}^2 + C_1)\,(\sigma_{\hat{x}_w}^2 + \sigma_{x_w}^2 + C_2)} 
\end{equation}
This window-based evaluation makes the SSIM above all sensitive to the mid frequencies, the structure and the contrast
at the window scale. The structure term captures a part of the high frequencies, but the index saturates near
$1$ and remains largely governed by the luminance agreement, that is, by the low frequencies. 

\paragraph{Wasserstein-1 distance ($\downarrow$):}\label{app:wass}
rather than confronting the pixels one by one, the Wasserstein distance compares the value distributions of the
two images. After having sorted separately the values of the
prediction, $\hat{x}_{(1)} \le \dots \le \hat{x}_{(M)}$, and those of the truth, $x_{(1)} \le \dots \le x_{(M)}$,
it equals the mean of the deviations between values of the same rank:
\begin{equation}
W_1(\hat{x}_{high}, x_{high}) = \frac{1}{M} \sum_{k=1}^{M} \big\lvert \hat{x}_{(k)} - x_{(k)} \big\rvert 
\end{equation}
It thus measures the statistical realism of the model, its ability to produce the right proportion of low,
medium and extreme values, independently of their position. It therefore carries no
spatial-frequency information, since it abstracts away the arrangement of the pixels. It nevertheless
reflects, indirectly, the loss of the high frequencies. A smoothing reduces the variance and compresses the tails of
the distribution, that is, the extremes, a deviation that this distance penalises.

\paragraph{Spectral error (PSD, $\downarrow$):}\label{app:spec}
this metric places itself directly in the frequency domain. The 2D Fourier transform of
each image is computed, $\hat{F} = \mathrm{FFT2}(\hat{x}_{high})$, from which the power spectral density is derived
$P = \lvert \hat{F} \rvert^2$, brought to logarithmic scale $L = 10\log_{10}(P + \epsilon)$ in order to balance
the weight of the low and high frequencies. The error is then the mean squared deviation between the
logarithmic spectra, the sum being over the spatial frequencies $k$: 
\begin{equation}
\mathrm{PSD}(\hat{x}_{high}, x_{high}) = \frac{1}{M} \sum_{k} \big( L_{\hat{x}}(k) - L_{x}(k) \big)^2 
\end{equation}
It is the only explicitly frequency-based metric. It
compares the energy present at each scale, from the low to the high frequencies. It therefore penalises the
high-frequency energy deficit that characterises a smoothed reconstruction.

\paragraph{Gradient error (Sobel, $\downarrow$):}\label{app:grad}
the gradient error judges the sharpness by comparing the edges of the two images. The spatial derivatives
are approximated by convolution with the Sobel filters $S_x$ and $S_y$. The gradient magnitude is
$\lVert \nabla u \rVert = \sqrt{(S_x * u)^2 + (S_y * u)^2}$. The metric is the MAE between the gradient
magnitudes of the prediction and of the truth:
\begin{equation}
\mathrm{GE}(\hat{x}_{high}, x_{high}) = \frac{1}{M} \sum_{p}
\big\lvert\, \lVert \nabla \hat{x}_{high} \rVert(p) - \lVert \nabla x_{high} \rVert(p) \,\big\rvert 
\end{equation}
The gradient being a high-pass operator, this metric targets the high frequencies, the edges and the
fine structures. It remains insensitive to the low frequencies, a constant shift not modifying the
derivatives. Unlike the spectral density, it however remains aligned. It requires that the edges
be reproduced at the right place and not only in the right quantity.

\subsection{Detailed Results}

Table~\ref{tab:bench_all_values} makes explicit the numerical values of Figure~\ref{fig:bench_grid}.
Tables~\ref{tab:mae} to~\ref{tab:grad} report the exact results 
on the test set for the five
metrics and the five methods U-Net, INR-SR, SE-OT, CLR and Interpolation, for different combinations
of climate variables. 

In each row, the best method is indicated in \textbf{bold} and the
second is \underline{underlined}.

\begin{table}[H]
  \caption{Numerical values of Figure~\ref{fig:bench_grid}: normalised scores (``best $=1$'')}
  \label{tab:bench_all_values}
  \centering
  \begin{tabular}{llccccc}
    \toprule
    Variable & Metric & U-Net & INR-SR & SE-OT & CLR & Interp \\
    \midrule
    \multirow{5}{*}{\texttt{tas}}
      & MAE   & \textbf{1.00} & 0.36 & \underline{0.49} & 0.43 & 0.25 \\
      & SSIM  & \textbf{1.00} & 0.24 & 0.64 & \underline{0.66} & 0.08 \\
      & Wass. & \textbf{1.00} & 0.17 & \underline{0.31} & 0.22 & 0.25 \\
      & Spec. & 0.49 & 0.43 & \textbf{1.00} & \underline{0.98} & 0.13 \\
      & Grad. & \textbf{1.00} & 0.42 & \underline{0.77} & 0.75 & 0.28 \\
    \midrule
    \multirow{5}{*}{\texttt{u10}}
      & MAE   & \textbf{1.00} & \underline{0.61} & 0.42 & 0.40 & 0.34 \\
      & SSIM  & \textbf{1.00} & \underline{0.47} & 0.31 & 0.32 & 0.22 \\
      & Wass. & \textbf{1.00} & \underline{0.37} & 0.23 & 0.18 & 0.29 \\
      & Spec. & 0.75 & 0.76 & \underline{0.98} & \textbf{1.00} & 0.25 \\
      & Grad. & \textbf{1.00} & \underline{0.70} & 0.62 & 0.63 & 0.39 \\
    \midrule
    \multirow{5}{*}{\texttt{v10}}
      & MAE   & \textbf{1.00} & \underline{0.63} & 0.44 & 0.42 & 0.38 \\
      & SSIM  & \textbf{1.00} & \underline{0.52} & 0.32 & 0.34 & 0.25 \\
      & Wass. & \textbf{1.00} & \underline{0.39} & 0.25 & 0.19 & 0.31 \\
      & Spec. & 0.69 & 0.62 & \textbf{1.00} & \underline{0.96} & 0.26 \\
      & Grad. & \textbf{1.00} & \underline{0.73} & 0.66 & 0.67 & 0.45 \\
    \midrule
    \multirow{5}{*}{\texttt{u10-v10}}
      & MAE   & \textbf{1.00} & \underline{0.66} & 0.36 & 0.35 & 0.34 \\
      & SSIM  & \textbf{1.00} & \underline{0.51} & 0.25 & 0.26 & 0.20 \\
      & Wass. & \textbf{1.00} & \underline{0.47} & 0.19 & 0.17 & 0.30 \\
      & Spec. & 0.73 & 0.69 & \underline{0.99} & \textbf{1.00} & 0.26 \\
      & Grad. & \textbf{1.00} & \underline{0.74} & 0.61 & 0.62 & 0.40 \\
    \midrule
    \multirow{5}{*}{\texttt{pr}}
      & MAE   & \textbf{1.00} & \underline{0.78} & 0.60 & 0.58 & 0.77 \\
      & SSIM  & \textbf{1.00} & \underline{0.72} & 0.45 & 0.46 & 0.67 \\
      & Wass. & \textbf{1.00} & 0.49 & 0.42 & 0.34 & \underline{0.61} \\
      & Spec. & 0.45 & 0.44 & \textbf{1.00} & \underline{0.95} & 0.28 \\
      & Grad. & \textbf{1.00} & 0.82 & 0.78 & 0.83 & \underline{0.84} \\
    \bottomrule
  \end{tabular}
\end{table}

\begin{table}[H]
  \caption{Normalised MAE, mean $\pm$ standard deviation - smaller is better.}
  \label{tab:mae}
  \centering
  \begin{tabular}{lccccc}
    \toprule
    Type & U-Net & INR-SR & SE-OT & CLR & Interp \\
    \midrule
    tas     & \textbf{0.040 $\pm$ 0.051} & 0.112 $\pm$ 0.127 & \underline{0.083 $\pm$ 0.074} & 0.094 $\pm$ 0.086 & 0.159 $\pm$ 0.202 \\
    u10     & \textbf{0.068 $\pm$ 0.081} & \underline{0.112 $\pm$ 0.127} & 0.164 $\pm$ 0.161 & 0.172 $\pm$ 0.172 & 0.200 $\pm$ 0.224 \\
    v10     & \textbf{0.079 $\pm$ 0.108} & \underline{0.126 $\pm$ 0.152} & 0.181 $\pm$ 0.180 & 0.188 $\pm$ 0.188 & 0.210 $\pm$ 0.251 \\
    u10-v10 & \textbf{0.070 $\pm$ 0.088} & \underline{0.106 $\pm$ 0.115} & 0.192 $\pm$ 0.189 & 0.198 $\pm$ 0.200 & 0.205 $\pm$ 0.238 \\
    pr      & \textbf{0.163 $\pm$ 0.387} & \underline{0.209 $\pm$ 0.467} & 0.273 $\pm$ 0.533 & 0.280 $\pm$ 0.498 & 0.213 $\pm$ 0.467 \\
    \bottomrule
  \end{tabular}
\end{table}

\begin{table}[H]
  \caption{SSIM - larger is better.}
  \label{tab:ssim}
  \centering
  \begin{tabular}{lccccc}
    \toprule
    Type & U-Net & INR-SR & SE-OT & CLR & Interp \\
    \midrule
    tas     & \textbf{0.987} & 0.945 & 0.979 & \underline{0.980} & 0.832 \\
    u10     & \textbf{0.945} & \underline{0.883} & 0.821 & 0.829 & 0.750 \\
    v10     & \textbf{0.939} & \underline{0.883} & 0.811 & 0.819 & 0.755 \\
    u10-v10 & \textbf{0.951} & \underline{0.904} & 0.804 & 0.810 & 0.761 \\
    pr      & \textbf{0.942} & \underline{0.920} & 0.871 & 0.875 & 0.914 \\
    \bottomrule
  \end{tabular}
\end{table}

\begin{table}[H]
  \caption{Wasserstein distance - smaller is better.}
  \label{tab:wass}
  \centering
  \begin{tabular}{lccccc}
    \toprule
    Type & U-Net & INR-SR & SE-OT & CLR & Interp \\
    \midrule
    tas     & \textbf{0.0819} & 0.478 & \underline{0.262} & 0.376 & 0.324 \\
    u10     & \textbf{0.0626} & \underline{0.170} & 0.270 & 0.342 & 0.219 \\
    v10     & \textbf{0.0642} & \underline{0.165} & 0.258 & 0.334 & 0.210 \\
    u10-v10 & \textbf{0.0649} & \underline{0.137} & 0.334 & 0.390 & 0.214 \\
    pr      & \textbf{0.000270} & 0.000553 & 0.000642 & 0.000790 & \underline{0.000446} \\
    \bottomrule
  \end{tabular}
\end{table}

\begin{table}[H]
  \caption{Spectral error - smaller is better.}
  \label{tab:spectral}
  \centering
  \begin{tabular}{lccccc}
    \toprule
    Type & U-Net & INR-SR & SE-OT & CLR & Interp \\
    \midrule
    tas     & 32.2 & 36.0 & \textbf{15.7} & \underline{16.0} & 117.8 \\
    u10     & 40.2 & 39.6 & \underline{30.8} & \textbf{30.2} & 119.1 \\
    v10     & 45.5 & 50.8 & \textbf{31.6} & \underline{33.0} & 123.1 \\
    u10-v10 & 43.6 & 46.5 & \underline{32.2} & \textbf{32.0} & 121.1 \\
    pr      & 209.6 & 211.4 & \textbf{93.6} & \underline{98.8} & 335.8 \\
    \bottomrule
  \end{tabular}
\end{table}

\begin{table}[H]
  \caption{Gradient error - smaller is better.}
  \label{tab:grad}
  \centering
  \begin{tabular}{lccccc}
    \toprule
    Type & U-Net & INR-SR & SE-OT & CLR & Interp \\
    \midrule
    tas     & \textbf{1.090} & 2.574 & \underline{1.417} & 1.445 & 3.895 \\
    u10     & \textbf{0.844} & \underline{1.210} & 1.368 & 1.344 & 2.192 \\
    v10     & \textbf{0.862} & \underline{1.179} & 1.302 & 1.279 & 1.900 \\
    u10-v10 & \textbf{0.818} & \underline{1.104} & 1.348 & 1.328 & 2.046 \\
    pr      & \textbf{0.00326} & 0.00397 & 0.00415 & 0.00394 & \underline{0.00389} \\
    \bottomrule
  \end{tabular}
\end{table}

\begin{table}[H]
  \caption{MAE (per method) and roughness $1-\rho_1$, relative to \texttt{tas}.}
  \label{tab:mae_rough_ratio}
  \centering
  \begin{tabular}{lcccccc}
    \toprule
     & \multicolumn{5}{c}{MAE ($\times$\,tas)} & Roughness ($\times$\,tas) \\
    \cmidrule(lr){2-6}
    Variable & U-Net & INR-SR & SE-OT & CLR & Interp & $1-\rho_1$ \\
    \midrule
    tas & 1.00 & 1.00 & 1.00 & 1.00 & 1.00 & 1.00 \\
    u10 & 1.69 & 1.00 & 1.98 & 1.83 & 1.26 & 1.72 \\
    v10 & 1.96 & 1.12 & 2.18 & 2.00 & 1.32 & 1.79 \\
    pr  & 4.04 & 1.86 & 3.30 & 2.98 & 1.35 & 4.11 \\
    \bottomrule
  \end{tabular}
\end{table}

\begin{table}[H]
    \centering
    \caption{Size and computation time of the models - example with tas}
    \label{tab:models_infos}
    \begin{tabular}{lcccc}
        \toprule
        \textbf{Model}             & \textbf{Parameters} & \textbf{Training} & \textbf{Inference}\\
                                    &                     &  \textbf{per epoch}   & \\
        \midrule
        \texttt{U-Net}              & \numprint{7993408}  & 14.83s                 & 2.00s \\
        \texttt{INR-SR}               & \numprint{2889217}  & 15.08s                 & 2.05s \\
        \texttt{SE-OT}               & 0                   & {\textemdash}         & 1.15s \\
        \texttt{CLR}                & \numprint{1671680}  & 0.74s                 & 3.00s \\
        \texttt{Interpolation}      & 0                   & {\textemdash}         & 0.15s \\
        \bottomrule
    \end{tabular}
\end{table}

\subsection{Reconstruction examples}

\begin{figure}[p]
  \centering
  \begin{subfigure}[t]{0.48\linewidth}\centering
    \includegraphics[width=\linewidth]{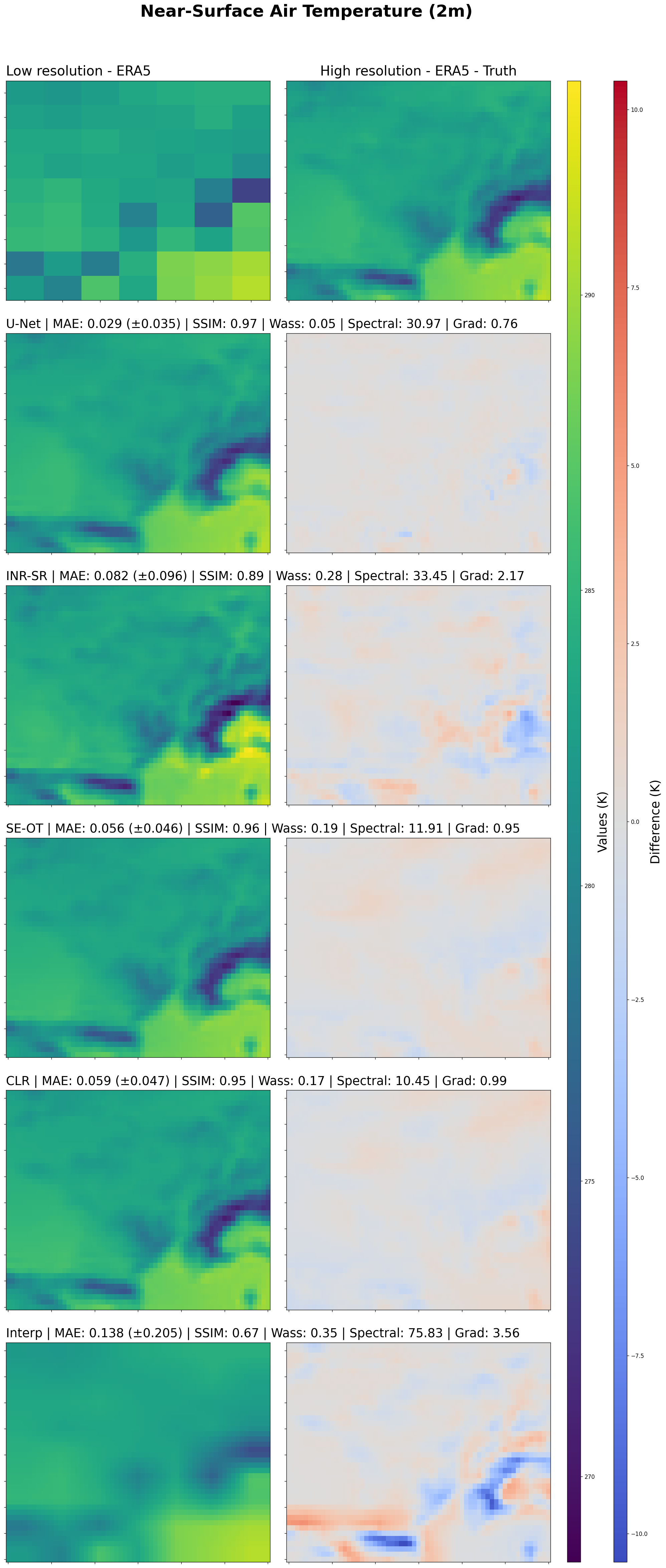}
    \caption{\texttt{tas}) U-Net obtains the best pointwise fidelity (MAE, SSIM), whereas SE-OT
    and CLR dominate the spectral error}
    \label{fig:verif_tas}
  \end{subfigure}\hfill
  \begin{subfigure}[t]{0.48\linewidth}\centering
    \includegraphics[width=\linewidth]{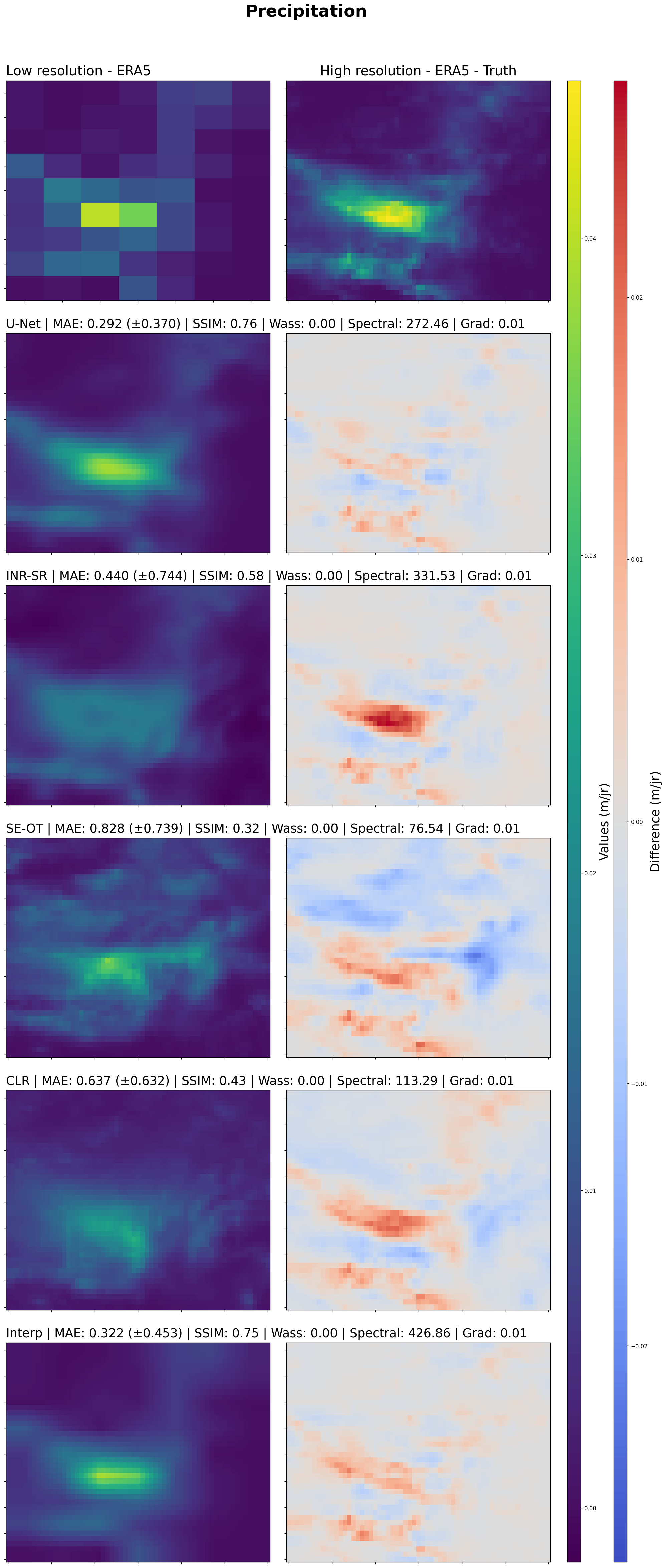}
    \caption{\texttt{pr} - sparse and intermittent field, the most difficult: the scores are
    compressed and the interpolation becomes competitive again on the pointwise metrics, because of the vast
    background of zero values}
    \label{fig:verif_pr}
  \end{subfigure}
  \caption{Reconstructions for \texttt{tas} and \texttt{pr}}
  \label{fig:verif_tas_pr}
\end{figure}

\begin{figure}[p]
  \centering
  \begin{subfigure}[t]{0.48\linewidth}\centering
    \includegraphics[width=\linewidth]{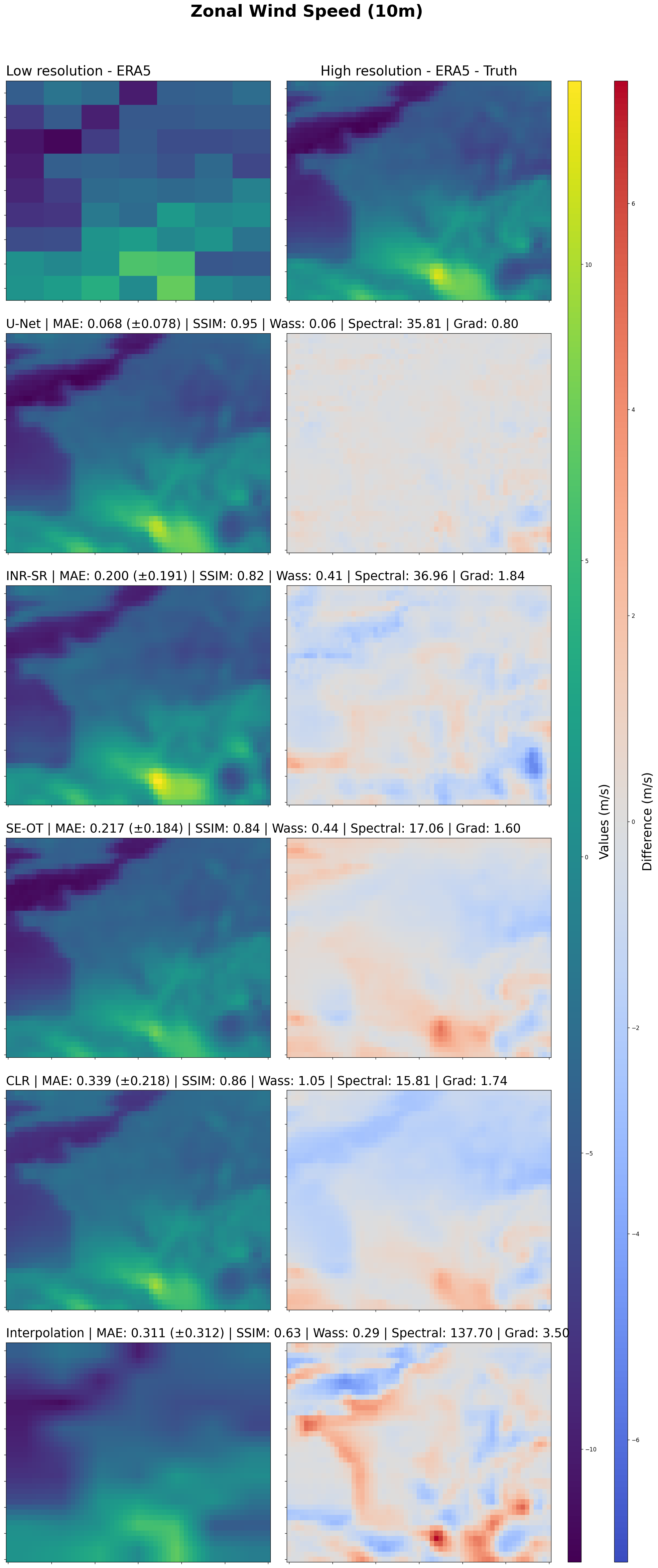}
    \caption{\texttt{u10}) - turbulent field: U-Net remains the best on the pointwise metrics and
    INR-SR establishes itself as a second}
    \label{fig:verif_u10}
  \end{subfigure}\hfill
  \begin{subfigure}[t]{0.48\linewidth}\centering
    \includegraphics[width=\linewidth]{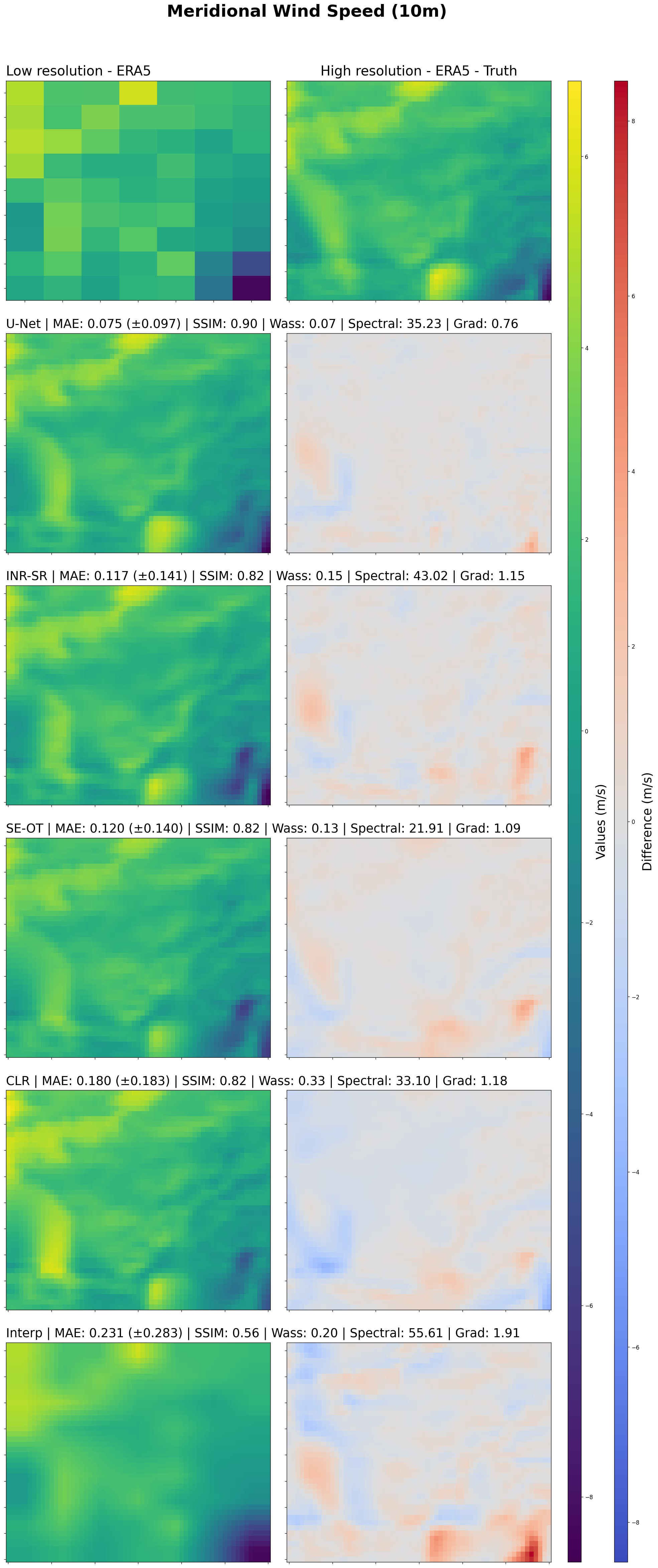}
    \caption{\texttt{v10} - turbulent field with a behaviour analogous to \texttt{u10}}
    \label{fig:verif_v10}
  \end{subfigure}
  \caption{Reconstructions for \texttt{u10} and \texttt{v10}}
  \label{fig:verif_u10_v10}
\end{figure}

\end{document}